\documentclass[final,1p,times]{elsarticle}
\usepackage{amssymb}
\usepackage{amsmath}
\usepackage{amsmath,amsfonts}
\usepackage{algorithmic}
\usepackage{algorithm}
\usepackage{array}
\usepackage[caption=false,font=normalsize,labelfont=sf,textfont=sf]{subfig}
\usepackage{textcomp}
\usepackage{stfloats}
\usepackage{url}
\usepackage{verbatim}
\usepackage{graphicx}
\usepackage{multirow}
\usepackage{placeins}

\begin{document}

\begin{frontmatter}

%% Title, authors and addresses

%% use the tnoteref command within \title for footnotes;
%% use the tnotetext command for theassociated footnote;
%% use the fnref command within \author or \affiliation for footnotes;
%% use the fntext command for theassociated footnote;
%% use the corref command within \author for corresponding author footnotes;
%% use the cortext command for theassociated footnote;
%% use the ead command for the email address,
%% and the form \ead[url] for the home page:
%% \title{Title\tnoteref{label1}}
%% \tnotetext[label1]{}
%% \author{Name\corref{cor1}\fnref{label2}}
%% \ead{email address}
%% \ead[url]{home page}
%% \fntext[label2]{}
%% \cortext[cor1]{}
%% \affiliation{organization={},
%%             addressline={},
%%             city={},
%%             postcode={},
%%             state={},
%%             country={}}
%% \fntext[label3]{}

\title{Enhancing 3D Object Detection in Autonomous Vehicles Based on Synthetic Virtual Environment Analysis}

%% use optional labels to link authors explicitly to addresses:
%% \author[label1,label2]{}
%% \affiliation[label1]{organization={},
%%             addressline={},
%%             city={},
%%             postcode={},
%%             state={},
%%             country={}}
%%
%% \affiliation[label2]{organization={},
%%             addressline={},
%%             city={},
%%             postcode={},
%%             state={},
%%             country={}}

\author[label1]{Vladislav Li}

\author[label2,label3]{Ilias Siniosoglou}

\author[label4]{Thomai Karamitsou}

\author[label4]{Anastasios Lytos}

\author[label5]{Ioannis D. Moscholios}

\author[label6]{Sotirios K. Goudos}

\author[label7]{Jyoti S. Banerjee}

\author[label2,label3]{Panagiotis Sarigiannidis}

\author[label1]{Vasileios Argyriou}

\affiliation[label1]{
  \institution{Kingston University}
  \department{Department of Networks and Digital Media}
  \city{Kingston upon Thames}
  \country{United Kingdom}
  \email{v.li@kingston.ac.uk, vasileios.argyriou@kingston.ac.uk}
}

\affiliation[label2]{
  \institution{University of Western Macedonia}
  \department{Department of Electrical and Computer Engineering}
  \city{Kozani}
  \country{Greece}
  \email{isiniosoglou@uowm.gr, psarigiannidis@uowm.gr}
}

\affiliation[label3]{
  \institution{MetaMind Innovations P.C.}
  \department{R\&D Department}
  \city{Kozani}
  \country{Greece}
  \email{isiniosoglou@metamind.gr, psarigiannidis@metamind.gr}
}

\affiliation[label4]{
  \institution{Sidroco Holdings Ltd.}
  \city{Nicosia}
  \country{Cyprus}
  \email{tkaramitsou@sidroco.com, alytos@sidroco.com}
}

\affiliation[label5]{
  \institution{University of Peloponnese}
  \department{Department of Informatics and Telecommunications}
  \city{Tripoli}
  \country{Greece}
  \email{idm@uop.gr}
}

\affiliation[label6]{
  \institution{Aristotle University of Thessaloniki}
  \department{Physics Department}
  \city{Thessaloniki}
  \country{Greece}
  \email{sgoudo@physics.auth.gr}
}

\affiliation[label7]{
  \institution{Bengal Institute of Technology}
  \city{Kolkata}
  \country{India}
  \email{tojyoti2001@yahoo.co.in}
}

%% Abstract
\begin{abstract}
%% Text of abstract
Autonomous Vehicles (AVs) use natural images and videos as input to understand the real world by overlaying and inferring digital elements, facilitating proactive detection in an effort to assure safety. A crucial aspect of this process is real-time, accurate object recognition through automatic scene analysis. While traditional methods primarily concentrate on 2D object detection, exploring 3D object detection, which involves projecting 3D bounding boxes into the three-dimensional environment, holds significance and can be notably enhanced using the AR ecosystem. This study examines an AI model's ability to deduce 3D bounding boxes in the context of real-time scene analysis while producing and evaluating the model's performance and processing time, in the virtual domain, which is then applied to AVs. This work also employs a synthetic dataset that includes artificially generated images mimicking various environmental, lighting, and spatiotemporal states. This evaluation is oriented in handling images featuring objects in diverse weather conditions, captured with varying camera settings. These variations pose more challenging detection and recognition scenarios, which the outcomes of this work can help achieve competitive results under most of the tested conditions.
\end{abstract}

%%Graphical abstract
% \begin{graphicalabstract}
% %\includegraphics{grabs}
% \end{graphicalabstract}

%%Research highlights
% \begin{highlights}
% \item Research highlight 1
% \item Research highlight 2
% \end{highlights}

%% Keywords
\begin{keyword}
%% keywords here, in the form: keyword \sep keyword
Augmented Reality\sep Object Detection\sep Scene Analysis\sep Scene Understanding\sep Object Recognition\sep Deep Learning\sep Feature Extraction.
%% PACS codes here, in the form: \PACS code \sep code

%% MSC codes here, in the form: \MSC code \sep code
%% or \MSC[2008] code \sep code (2000 is the default)

\end{keyword}

\end{frontmatter}

%% Add \usepackage{lineno} before \begin{document} and uncomment 
%% following line to enable line numbers
%% \linenumbers

%% main text
%%

% DONE
\section{Introduction}

In the world of autonomous driving, scene analysis and comprehension serve as the bedrock upon which the vehicle's perception and interaction with its surroundings are built \cite{pavel2022vision} \cite{xiong2021augmented}. Advanced computer vision and machine learning (ML) algorithms are employed to process vast amounts of data gathered from the vehicle's sensors, such as cameras, LiDAR, and radar. The recognition and interpretation of the ever-changing surroundings allow the vehicle to make informed choices about navigation, safety, and interactions with other road users.

For autonomous vehicles, scene analysis and comprehension play an important role. This includes a wide range of applications such as detecting other vehicles sharing the road, recognizing traffic signs, as well as detecting pedestrians, potential hazards, etc. This deeper understanding is instrumental in making autonomous decisions while integrating the augmented environment onto the vehicle's display systems like heads-up displays (HUDs). This extends to visual scene analysis which is the cornerstone of vehicle environment perception and interaction using advanced computer vision machine learning (ML) algorithms for controlling large amounts of data collected from sensors, such as cameras, LiDAR, and radar. The recognition and interpretation of the ever-changing surroundings allow the vehicle to make informed choices about navigation, safety, and interactions with other road users.

In order to drive safely and effectively, for example, the car must be able to recognize and identify other vehicles, as well as their position and relative speed. Equally important is the recognition of traffic signs and signals, ensuring compliance with traffic regulations and the seamless flow of traffic. Moreover, the accurate detection of pedestrians, cyclists, and potential obstacles is indispensable for avoiding accidents and ensuring the safety of all road users.

In the last decades, advances in computer vision have fostered the design and implementation of object recognition methods, increasing computational performance and lowering process time \cite{zou2023object}. These technologies enable the vehicle's onboard computer systems to continuously learn and adapt, improving their ability to recognize and respond to an ever-evolving array of environmental conditions. An important milestone is that in the optimisation phase of such applications, the evaluation of AV image cognition systems can be performed in the virtual and augmented reality domains, utilising the same environment that is also used in virtual applications, like game development engines. As a result, current scene analysis technologies based on object recognition use complex computer vision techniques to detect and track objects in the real world. Examples of such technologies include the You Only Look Once (YOLO) model \cite{anderson2019feasibility}, homomorphic filtering and Haar markers \cite{gomes2018augmented} and the Single Shot Detector \cite{dimitropoulos2021operator}. The use of Convolutional Neural Networks (CNNs) and Deep Learning (DL) led to faster and more accurate detection processes \cite{zhao2019object}. However, the AR experience could be improved by projecting 3D objects into the augmented reality space surrounding the user inferred from the real environment.

% Main contribution.
The aim of this study is to analyse a novel 3D solution that evaluates the performance of the 3D bounding box prediction in various conditions. This work proposes a novel architecture to efficiently produce 3D bounding boxes, superimposed onto the multivariate spatiotemporal view that technologies like advanced AR and AV cognition systems employ to perceive the three-dimentional environment. The produced system is evaluated on data further augmented by producing a synthetic dataset that encapsulates a variety of possible environmental conditions, like, camera view, lighting, weather, and sensor readings. Finally, this study evaluates the proposed architecture with other benchmark methods, providing a comparative dimension. The main contributions of this work are as follows:

\begin{itemize}
    \item A multimodal architecture for efficient object detection and localisation for real-time scene analysis
    \item A methodology for predicting 3D bounding boxes on the three-dimensional environment, extrapolated from 2D images
    \item A Novel Synthetic Image dataset for object detection in AV applications with VR scene augmentation
    \item A comparative study of the efficacy and efficiency of the developed methodology against state-of-the-art techniques
\end{itemize}

% The navigation.
The rest of this paper is organised as follows: \ref{Overview of Previous Work} provides an overview of related work. \ref{Methodology} describes the proposed architecture. \ref{Evaluation} presents results obtained using a novel synthetic image dataset. Finally, \ref{Conclusion} concludes this work.

\section{Overview of Previous Work}
\label{Overview of Previous Work}

\subsection{Region-based Feature Extraction Algorithms}
An AR app identifies objects in the real world using ML and computer vision techniques with the goal of overlaying virtual objects in real-time. In recent years, the use of deep CNNs \cite{lecun1998gradient} has greatly enhanced the performance and accuracy of object detection and recognition in computer vision. In 2014, Girshick et al. introduced the Regions with CNN features (RCNN) method for object detection \cite{girshick2014rich}. This approach involved first identifying potential object boxes through selective search and then rescaling each box to a fixed-size image for input into a CNN model trained on AlexNet \cite{viola2001rapid} for feature extraction. The object was then detected using a linear SVM classifier, resulting in a significant improvement in mean Average Precision compared to previous methods, but also had a significant drawback of slow detection speed.

In 2014, Girishick et al. introduced the "Regions with CNN features" (RCNN) method for the purpose of object detection, as documented in their seminal work \cite{girshick2014rich}. This pioneering approach signified a significant breakthrough in the realm of computer vision, particularly concerning the enhancement of object detection accuracy. The RCNN methodology employed a dual-stage process. Firstly, it commenced with the utilization of "selective search" to identify prospective object boxes within an image. Selective search effectively partitioned the image into multiple regions or proposals that were posited as likely candidates harboring objects. These regions were thereby considered as candidate boxes for potential object localization. 

Subsequently, the next steps in the RCNN procedure entailed the resizing of the aforementioned candidate bounding boxes to fit into fixed-size images, rendering them ready for analysis. These standardized images are then subjected to a CNN-based processing, specifically pre-trained on the AlexNet model \cite{viola2001rapid}. The principal role of this CNN was to perform feature extraction in order to discern and capture highly distinctive features of the object in question. Upon feature extraction, the final step of the RCNN methodology involved the employment of a linear Support Vector Machine (SVM) classifier. The SVM classifier was instrumental in effecting classification of the extracted features, thereby ascertaining the presence or absence of a given object within the candidate box. This classification process was the basis of object identification and localization.

The outcomes of the RCNN approach bore substantial significance. It led to a marked augmentation in the mean Average Precision metric, a pivotal gauge of the efficacy and precision of object detection algorithms. Effectively, it surpassed antecedent methods in its competence to identify objects within images, marking a substantial progression in the arena of computer vision.

Nevertheless, it is worth acknowledging that the RCNN method suffered from a comparatively lengthy detection timeframe which can majorly impact the overall performance. Its sequential operations, such as selective search, CNN-based feature extraction, and SVM classification, made it very computationally intensive and took a long time to process, which limited its usefulness in situations where real-time object detection was needed.

So, while the RCNN approach made it easier to find objects, it required a lot of computing power and time, which meant that more research had to be done to make it work faster. Regardless, its creation was a major turning point in the history of object detection algorithms. It paved the way for later innovations and sped up progress in areas like robotics, autonomous vehicle systems, and many types of computer vision applications.

In an effort to tackle the persistent challenge of slow detection speed in object recognition and localization, He et al. presented the Spatial Pyramid Pooling Network (SPPNet) as an innovative solution in their seminal work \cite{he2015spatial}. This architectural paradigm marked a notable milestone in the evolution of computer vision, offering a profound remedy to a long-standing predicament in the field.

The basis of the SPPNet's success lay in its strategic incorporation of a Spatial Pyramid Pooling (SPP) layer, a pivotal component that revolutionized the object detection process. The distinctive feature of this SPP layer was its ability to generate a fixed-length representation that remained invariant to alterations in image size and scale. This attribute had far-reaching implications, particularly in terms of mitigating overfitting issues that had previously plagued object recognition systems. After this initial feature extraction step, the SPPNet employed a sub-region pooling mechanism. This operation entailed dividing the image into spatial bins, enabling the aggregation of features from each bin to create fixed-length representations that were conducive for detector training.

One of the most notable outcomes of this innovative approach was a remarkable acceleration in processing speed, especially during testing. The SPPNet method proved to be a significant leap forward, with testing times ranging from 24 to 102 times faster than the previously established RCNN approach. This acceleration in speed held profound implications for real-time and time-sensitive applications, particularly in contexts like autonomous vehicles, robotics, and augmented reality.

\begin{figure}
 \centering
  \includegraphics[scale=0.65]{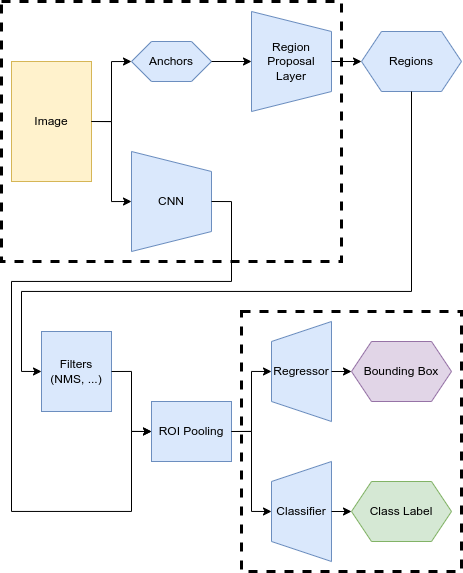}
 \caption{FRRCNN architecture.}
 \label{fig:abstract-frrcnn-architecture}
\end{figure}

In 2015, Girishick improved the previous two architectures with Fast RCNN \cite{girshick2015fast}. This network trains both a detector and a bounding box regression simultaneously with the same configuration. However, the speed limitation persisted. The same year, Ren et al. introduced the Faster RCNN detector \cite{ren2015faster}, which was the first deep learning detector to almost achieve real-time detection through end-to-end training. This architecture employed the Region Proposal Network (RPN) to speed up the detection process, and several variants have been proposed since then to reduce computational redundancy \cite{dai2016r}, \cite{li2017light}, \cite{lin2017feature}. In particular, Cao et al. (2020) \cite{cao2020d2det} introduced the D2Det method based on the Faster R-CNN framework, which processes Region of Interest (ROI) features through two stages: high-density local regression and discriminant ROI pooling. The method replaces the Faster RCNN offset regression with a local dense regression block. Girishick furthermore introduced an enhancement to the existing architectural paradigms in the form of the Fast RCNN \cite{girshick2015fast}. This novel network configuration entailed the simultaneous training of both an object detector and a bounding box regression component, all within the same unified architecture. However, it is noteworthy that the issue of computational speed constraints persisted despite this development.

The Fast RCNN model builds upon the existing state-of-the-art, enhancing efficiency. It exhibites the capability to concurrently train two fundamental components within the same system, i) an object detector and ii) a bounding box regression module, incorporated under the same framework. This integrated approach was a significant stride towards a more streamlined and coherent training process. Nevertheless, the overarching challenge of computational speed constraints persisted as an obstinate issue in the field.

In the same time, Ren et al. introduced the Faster RCNN detector \cite{ren2015faster}, a groundbreaking endeavor that charted a course toward the realization of real-time object detection through the prism of end-to-end training. The Faster RCNN architecture marked a seminal turning point in the pursuit of swifter detection capabilities. At its core, it introduced the Region Proposal Network (RPN), a component specifically designed to expedite the object detection process. The RPN's mandate involved the generation of region proposals, a facet that greatly enhanced the network's adeptness in efficiently discerning objects within complex scenes.

The introduction of the Faster RCNN model had an indelible impact on the landscape of computer vision. It not only ushered in the possibility of near real-time object detection but also spurred a wave of innovative architectural variants. These variations, with an overarching focus on curtailing computational redundancy \cite{dai2016r}, \cite{li2017light}, \cite{lin2017feature}, explored diverse avenues to further amplify the velocity and efficiency of object detection while preserving precision.

Among these progressive adaptations, the D2Det method, introduced by Cao et al. in 2020 \cite{cao2020d2det}, stands out as an exemplar of innovation based on the Faster R-CNN framework. The D2Det method harnesses a sophisticated two-stage process for handling Region of Interest (ROI) features. In the initial phase, high-density local regression is employed to finetune the localization of objects, infusing a heightened degree of precision into the detection process. Subsequently, in the second stage, a discriminant ROI pooling mechanism extracts distinctive features from the ROIs. Notably, D2Det departs from the Faster RCNN's offset regression by adopting a local dense regression block, thus augmenting the precision and robustness of the object detection process.

The progress made by researchers and the path from Fast RCNN to Faster RCNN and beyond show that the goal of real-time object recognition is still being worked on and will continue to get better. There is a chance that these advances will change many areas, such as autonomous systems, surveillance, robots, and augmented reality. At the cutting edge of progress in computer vision and deep learning is the never-ending search for faster, more accurate, and more efficient ways to find objects.

The methodologies discussed above fall under the classification of two-stage detectors due to their characteristic two-step process: initially generating regions of interest (ROIs) and subsequently executing detection and recognition. In 2016, Joseph et al. introduced a noteworthy departure from this convention, presenting a one-stage detector known as You Only Look Once (YOLO) \cite{redmon2016you}. YOLO epitomized a pioneering paradigm shift in the realm of object detection, manifesting as a single network architecture capable of processing the entirety of an image within a solitary step, which resulted in substantially expedited processing times.

\subsection{Segmentation-based Feature Extraction}

\begin{figure}
 \centering
 \includegraphics[scale=0.65]{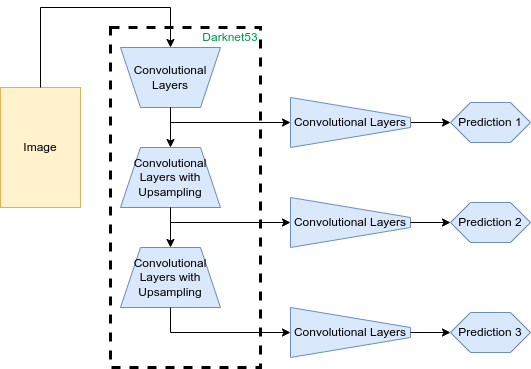}
 \caption{YOLO architecture.}
 \label{fig:abstract-yolov3-architecture}
\end{figure}

The YOLO methodology operates by segmenting the image into distinct regions and concurrently predicting bounding boxes for each of these regions. This one-step processing paradigm marked a significant departure from the multi-step procedures of its two-stage counterparts.

YOLO was a game-changer in the field, representing a revolutionary approach to object detection. Unlike its two-stage counterparts, YOLO employed a single neural network architecture, capable of processing an entire image in a solitary pass. This unique design offered a significant advantage in terms of processing speed, effectively reducing detection times. The core principle underlying YOLO's functionality involved the division of the image into discrete regions, with the network making concurrent predictions for bounding boxes associated with each region. This approach eliminated the need for sequential processing, offering a substantial boost in efficiency. In the subsequent years, YOLO underwent iterations with the introduction of YOLO v2 and v3, aimed at enhancing prediction accuracy \cite{redmon2017yolo9000}, \cite{redmon2018yolov3}, while subsequent versions, v5 through v8 focus on prediction efficiency, accuracy, speed and deployment optimisation. 

To enhance the accuracy of bounding box localization, $DIoU$ loss is employed due to its demonstrated improvement in performance when used with the YOLO algorithm \cite{Wang2022}. $DIoU$ represents an advancement of the IoU metric (\ref{eq:iou_metric}), specifically targeting the optimization of bounding box predictions. 

\begin{equation}
IoU = \frac{\textrm{Area of Overlap}}{\textrm{Area of Union}} = IoU(B_p,B_r) = \frac{B_p \cap B_r}{B_p \cup B_r}
\label{eq:iou_metric}
\end{equation}
and the distance
\begin{equation}
Loss_{IoU} = 1 - IoU
\label{eq:loss_iou_metric}
\end{equation}

In this context, $B_p$ and $B_r$ represent the predicted and actual bounding boxes, respectively. The term $DIoU$ improves upon $IoU$ by factoring in the square of the diagonal $d_B$ of the smallest bounding box $B_o$ that encompasses both $B_p$ and $B_r$. Therefore, the equation is as follows:

\begin{equation}
DIoU = IoU - \frac{\sqrt{(B_{p}^{2}) - (B_{r}^{2})}^2}{d_{B}^{2}}
\label{eq:diou_metric}
\end{equation}
and the resulting loss function.

\begin{equation}
L_{DIoU} = 1 - DIoU = 1 -  IoU - \frac{\sqrt{(B_{p}^{2}) - (B_{r}^{2})}^2}{d_{B}^{2}}
\label{eq:loss_diou_metric}
\end{equation}

This function, by addressing the issue of non-intersecting bounding boxes in terms of $IoU$, aids in accelerating the model's convergence.

While YOLO excelled in terms of speed, it encountered challenges related to localization accuracy. This trade-off spurred further research efforts to fine-tune the model. To redress this trade-off and enhance the localization accuracy, Liu et al. introduced the Single Shot MultiBox Detector (SSD) in 2016 \cite{liu2016ssd}. The SSD method was different from the one-stage paradigm because it used both multi-reference and multi-resolution detection strategies. This made it possible to find objects at different sizes across different network layers. This architecture can accommodate objects of diverse sizes and magnitudes within the image, mitigating the aforementioned accuracy compromise.

In 2018, Lin et al. presented RetinaNet \cite{lin2017focal}, marking a significant advancement in one-stage object detection. The key innovation within RetinaNet was the introduction of a novel loss function termed "focal loss." This loss function, which differs from the cross-entropy loss, was created to give more attention to instances that kept getting incorrectly labeled during the training process. This heightened attention to challenging examples during training resulted in an enhanced level of prediction accuracy, outstripping the performance of its one-stage counterparts.

\subsection{Anchor-free Inference}

In contemporary developments within the domain of object detection, there is a noteworthy shift towards anchor-free methodologies \cite{Xinyu2023}. These novel approaches, in contrast to conventional techniques, emphasize the inference of bounding box corners, rather than reliance on pre-defined bounding boxes. A prominent exemplar of this trend is the CenterNet, an innovative framework introduced by Zhou et al. \cite{zhou2019objects}. Notably, CenterNet has distinguished itself as a state-of-the-art solution for 3D Lidar-based detection and tracking, showcasing its versatility in diverse applications.

CenterNet can be perceived as an evolution of the CornerNet, another anchor-free approach to bounding box detection that represents objects as pairs of keypoints, specifically the top-left and bottom-right corners. These corner keypoints are extracted through a technique known as corner pooling, which was introduced by the same authors \cite{law2018cornernet}. A critical stride in the advancement from CornerNet to CenterNet was the introduction of a central keypoint, a concept that facilitated the association of corner keypoints with objects depicted in images. This novel approach has demonstrated superior performance compared to conventional anchor-based solutions, such as Faster RCNN and YOLO, marking a significant advancement in object detection.

Continuing the trajectory of innovation, in 2020, Perez-Rua and colleagues introduced the OpeN-ended Center nEt (ONCE) \cite{Perez_Rua_2020_CVPR}. ONCE improved CenterNet's abilities by letting it find objects from classes where there were not many examples in its training dataset. This is an impressive achievement that could be useful in situations involving many types of objects.

Additionally, object detection techniques have begun to explore the capabilities of transformers, as paved by the DEtection TRansformer (DETR) method introduced by Carion et al. \cite{carion2020end}. This exploration leverages the advantages of transformer architectures, which have gained prominence in natural language processing, and integrates them into the object detection domain. What sets DETR apart is its simplicity, coupled with performance that rivals other sophisticated detection techniques employed in the field. Subsequently, Zhu et al.\cite{zhu2021deformable} proposed the Deformable DETR system. This system builds on the specific objective detecting small objects. This enhancement aimed to achieve state-of-the-art performance, underscoring the commitment of the scientific community to continuously refine and advance object detection methodologies to meet the evolving demands of real-world applications.

% Text is done, diagrams are done, NEED proof-reading.
\section{Methodology}
\label{Methodology}

This section delves into the methodology, presented in this work, for detecting 3D objects utilizing ML algorithms and the CenternNet architecture. The methodology employs offsets to refine corner positions and heatmaps for predicted key-pairs and corners, in relation to their corresponding object categories. An Anchor-free module produces heatmaps, where embedding methods establish significant connections between anticipated object classes and predicted corner values. The methodology further investigates the 3D bounding box regression process, which is leveraged to approximate attributes such as i) depth, ii) dimensions, and iii) orientation. The employed ML methods and algorithms enhance these to enable the thorough and precise detection of 3D objects in the respective scene. The following methodology plays a critical role in attaining accurate and comprehensive object detection.

\subsection{Methodology Overview}

The proposed architecture incorporates a multitude of smaller components. The first component predicts a 2D bounding box common to the standard object detection task. It achieves it by generating and processing Heatmaps, Embeddings, and Offsets \cite{law2018cornernet}. The three outputs are then further processed in Cascade Corner Pooling and Centre Pooling \cite{duan2019centernet} components which infer the final positions of the 2D bounding box. The next component then estimates depth using a Multi-Scale Deep Network \cite{eigen2014depth}, in two stages, the first stage collects global information, and, the second stage refines the global information to produce a more precise prediction. The next component estimates the 3D dimensions and orientation of the object using CNNs. The 3D dimensions are regressed directly from feature map outputs using fully-connected layers, however, the orientation is formulated as a classification task and regressed accordingly using the hybrid discrete-continuous MultiBin loss \cite{mousavian20173d}. The final 3D bounding box is constrained by the predicted 2D bounding box with an assumption that the 2D bounding box has been trained to match the position of the 3D bounding box.

\begin{figure}[!b]
    \centering
    \includegraphics[width=0.65\textwidth]{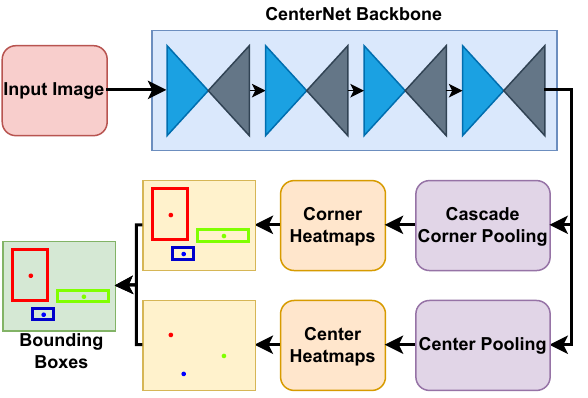}
    \caption{CenterNet architecture.}
    \label{fig:6}
\end{figure}

\subsection{Cascade Corner \& Center Pooling Spatial Localisation}

% A paragraph about the CenterNet's Cascade Corner Pooling and Center Pooling contribution.
The methodology presented in this study, depicted in Figure \ref{fig:6}, is fundamentally grounded in an anchor-free approach. Within this framework, keypoint descriptors play a fundamental role in representing crucial elements of object detection, encompassing essential entities such as the top-left, bottom-right, and center points. This method places a strong emphasis on the precise determination of both corner and center keypoints, a task of utmost significance, as it significantly influences the accuracy and reliability of object localization and recognition.

The basis of this approach lies in the meticulous discernment and characterization of keypoint descriptors and which serve as distinctive markers of object structure and spatial attributes. These descriptors capture not only the spatial coordinates but also the semantic information associated with the objects in question. This careful attention to detail contributes to the heightened accuracy and reliability of tasks related to object localization and recognition. These capabilities are of particular importance in various domains and including autonomous systems and surveillance and and computer vision research and where the ability to accurately and consistently identify and locate objects is a central requirement.

To confirm the corner keypoints of the bounding box, the methodology includes an AI model prepared with a Cascade Corner Pooling module. This module is chargeable for calculating the most summed response alongside the limits of the function map. Moreover, it extends its scope of analysis to encompass the internal instructions inside the function map. This approach showcases the stableness and robustness of the method, mainly in the face of function-stage noises and variations. The Cascade Corner Pooling module represents a complicated architectural innovation, optimized to provide a comprehensive and noise-tolerant mechanism for an appropriate localization of corner keypoints.

Subsequently, the methodology employs a dedicated AI model to infer the center keypoints. This task is facilitated by the addition of a Center Pooling module. This module calculates the maximum summed response along both horizontal and vertical directions within the feature maps. This bidirectional analysis is instrumental in pinpointing the central keypoints of objects, a task that is important for more accurate object localization and recognition.

The idea behind this methodology is based in the work of Duan et al. \cite{duan2019centernet} that defines the utilization of keypoint descriptors, coupled with data-driven methodologies for determining corner and center keypoints. This method is characterized by its robustness, accuracy, and adaptability, making it well-suited for a wide variety of applications, including but not limited to autonomous vehicle systems, robotics, and computer vision research. 

\subsection{3D Bounding Box Inference}

% A paragraph about the CornerNet's Heatmap, Embedding, and Offsets contribution.
The determination of keypoints within this framework is achieved through a systematic process that entails the utilization of Heatmaps derived from a set of feature maps generated by the Cascade Corner Pooling and Center Pooling modules. These Heatmaps serve as a representation of the approximate positions of keypoint entities, which are classified into three distinct categories: top-left, bottom-right, and center points. Notably, the Heatmaps are configured with a dimensionality of $C$ channels, where $C$ corresponds to the number of object classes under consideration. Additionally, they possess dimensions mirroring those of the input image, denoted as $H \times W$, with $H$ denoting the image's height and $W$ representing its width.

One of the main keypoints in this methodology is the generation of Associative Embeddings, which serve as a means to establish a link between individual keypoints and their respective object classes. These Embeddings $E={e_1,e_2,...,e_n}$ are computed by the AI model and disregard the need for a "ground-truth" label. Instead, their significance lies in the relative differences, which facilitate the grouping of object detections based on their associated Embeddings. Each detected embedding $e_i$ generated by the AI network is accompanied by a numerical value, denoted as a "tag" $t_i$, which plays an instrumental role in the subsequent grouping of detections. The premise is that detections with similar tags should be effectively clustered together in a respective tag cluster $C_i$. To break this down, let's assume a set of embedding detection $D$ incorporating detections $d_1,d_2,...,d_n$ where $D={d_1,d_2,...,d_n}$, produced by the model. Each detection $d_i$ is associated with an embedding $e_i$ and a numerical tag $t_i$, producing an Associated Embedding $E(di)\rightarrow(e_i,t_i)$, where $E$ is the embedding function. The produced tags are clustered assuming a clustering function $S(t_i,t_j)$, highlighting if a set of detections are similar when $S$ exceeds a threshold $\theta$. This relationship can be summarised as shown in Eq. \ref{embedding_corellation}.

\begin{equation}
\begin{aligned}
\forall d_i, d_j \in D, & \\
\text{ if } S(t_i, t_j) > \theta, & \\
\text{ then } (d_i, d_j) \in C_l \subseteq C(D)
\label{embedding_corellation}
\end{aligned}
\end{equation}

The output of the model necessitates certain adjustments to optimize the fit of the predicted object to the actual object within the image. In response to this requirement, Offsets are introduced to facilitate these adjustments. Let's assume the set of predicted keypoint locations in the featuremap $P_k={(x_p, y_p)}$ and a set of offset vectors $O = {(o_x, o_y)}$ with $o_x$ and $o_y$ each offset vector corresponding to a unique predicted keypoint $(x_p, y_p)$ in $P_k$. Each Offset map serves as a spatial mapping of keypoint locations within the feature map space to their corresponding positions on the input image, conveyed in pixel coordinates. This process aids in predicting a set of actual keypoint locations $A_k={(x_a, y_a)}$ in the resulting image following a correlation of $A_{ki} = P_{ki} + O_i$ for each $i$-th keypoint adjusted by its corresponding offset to align with the actual keypoint location in the image. The application of Offsets represents an essential mechanism for fine-tuning the localization of keypoints and enhancing the overall precision of object detection within the framework.

\begin{figure*}[]
    \centering
    \includegraphics[width=\textwidth]{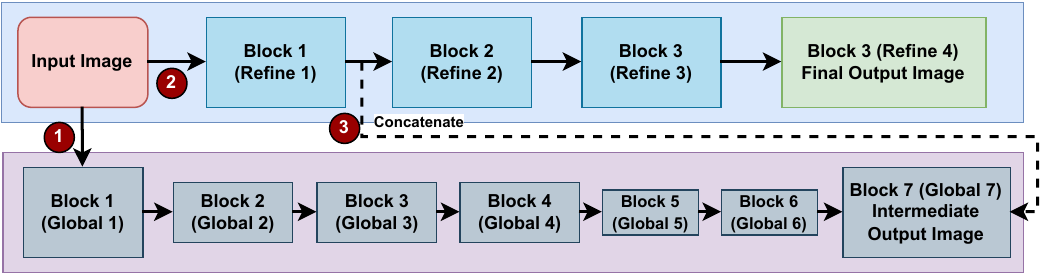}
    \caption{Abstract representation of the Multi-Scale Deep Network}
    \label{fig:1}
\end{figure*}

\begin{figure}
 \begin{center}   
  \includegraphics[scale=0.05]{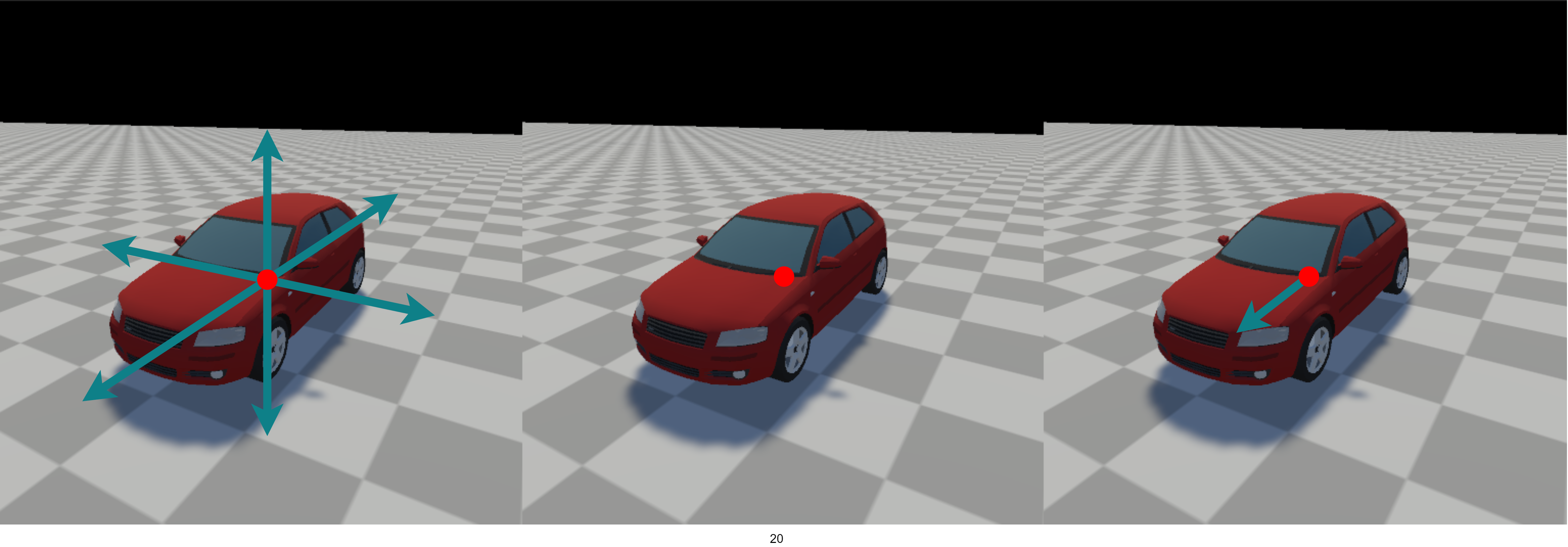}
 \end{center}
 \caption{The network output for 3D object detection. From left to right: 3D dimensions (metres), depth (metres), orientation (degrees).}
 \label{fig:7}
\end{figure}

% A paragraph about the Eigen's Depth contribution.
To achieve the successful detection of 3D objects, the methodology entails the prediction of center keypoints, incorporating additional information encompassing depth, 3D dimensions, and orientation, as visually depicted in Figure \ref{fig:7}. The depth component is a transformed output, derived from Eigen et al.'s approach \cite{eigen2014depth}, and it is presented as an additional scalar value associated with each center keypoint.

The depth prediction mechanism comprises two principal modules, as illustrated in Figure \ref{fig:1}. The initial component, the Global Coarse-Scale Network \cite{eigen2014depth}, takes the input image and endeavors to predict the depth of the entire scene at a global scale. This global-level prediction serves as the foundation for subsequent refinements. The refinement process is carried out by the Local Fine-Scale Network \cite{eigen2014depth}, which receives the output from the Global Coarse-Scale Network and fine-tunes the initial coarse predictions. This fine-tuning process is vital for aligning the depth predictions with local details, including the edges of objects and walls.

In assessing the quality of the depth predictions, the methodology employs a Scale Invariant Error metric (SIE), which can be seen in Eq. \ref{SIE}.

\begin{equation}
SIE = \frac{1}{n} \sum_{i=1}^{n} \left( \log d_i - \log \hat{d_i} - \frac{1}{n} \sum_{j=1}^{n} (\log d_j - \log \hat{d_j}) \right)^2
\label{SIE}
\end{equation}

Where $n$ is the number of data points, $d_i$ is the true value for the $i$-th data point and $\hat{d_i}$ is the predicted value for the $i$-th data point. This metric computes the per-pixel differences between the predicted depth map and the ground truth. Notably, the calculations are performed in a logarithmic space, a choice made to address the potential issue of the average scale of the scene influencing the error measurements. This approach ensures a more robust and scale-invariant assessment of the quality of the depth predictions, ultimately contributing to the accuracy of the 3D object detection process.

% A paragraph about the Mousavian's 3D Dimension and Orientation contribution.
The representation of a 3D bounding box in this context relies on three primary parameters, specifically the bounding box center, dimensions, and orientation. The center of the 3D bounding box is characterized by a set of three 3D coordinates, denoted as $x$, $y$, and $z$ (where the center $B_c=(x,y,z)$. The dimensions of the 3D bounding box are governed by an additional triad of attributes, namely width $w$, height $h$, and length $l$, measured in meters, represented as $B_d=(w,h,l)$. These dimensions are directly regressed to their respective attributes through the application of a straightforward loss function measuring the distance of the regressed samples, like Mean Square Error, for the three-dimensional aspect of the samples, Eq. \ref{MSE_3d}. The orientation of the 3D bounding box is defined by another set of three attributes, encompassing azimuth, elevation, and roll angles in degrees $B_o=(\theta_{\text{azimuth}}, \theta_{\text{elevation}}, \theta_{\text{roll}})$.

\begin{equation}
L_{MSE} = \frac{1}{n} \sum_{i=1}^{n} \left[ (w_i - \hat{w_i})^2 + (h_i - \hat{h_i})^2 + (l_i - \hat{l_i})^2 \right]
\label{MSE_3d}
\end{equation}

The derivation of the three 3D coordinates is realized through the utilization of the MultiBin architecture \cite{mousavian20173d}. In this architecture, each angle is treated as a distinct class, effectively framing the orientation prediction as a classification task. To account for the angular relationships between classes, the MultiBin architecture incorporates the computation of small offsets using trigonometric functions, specifically sine and cosine, applied to the angles. The outcome of this module encompasses three values for each class: the confidence associated with the class, the cosine difference of the angle, and the sine difference of the angle. Specifically, the angles are divided into \( N \) bins, each associated with a class \( C \) and bin \( i \), calculating a confidence score \( \text{Conf}_{C,i} \), and trigonometric offsets, \( \cos(\Delta \theta_{C,i}) \) and \( \sin(\Delta \theta_{C,i}) \), where \( \Delta \theta_{C,i} = \theta - \theta_{\text{center}, i} \). The final angle \( \hat{\theta}_{C} \) is estimated by combining these values, weighted by confidence scores, offering precise angular orientation predictions.

For the successful projection of a 3D bounding box onto a 2D image, the calculations necessitate the availability of a camera intrinsic matrix. This matrix plays an important role in ensuring the accurate alignment of the 3D bounding box with the 2D image. Furthermore, to enhance the precision and reliability of the 3D bounding box, it is constrained through the utilization of the 2D bounding box. This constraint mechanism contributes to the refinement of the 3D bounding box and enhances the accuracy of the overall object detection process.

\subsection{Parameters}

\begin{figure*}
 \begin{center} 
  \includegraphics[width=\textwidth]{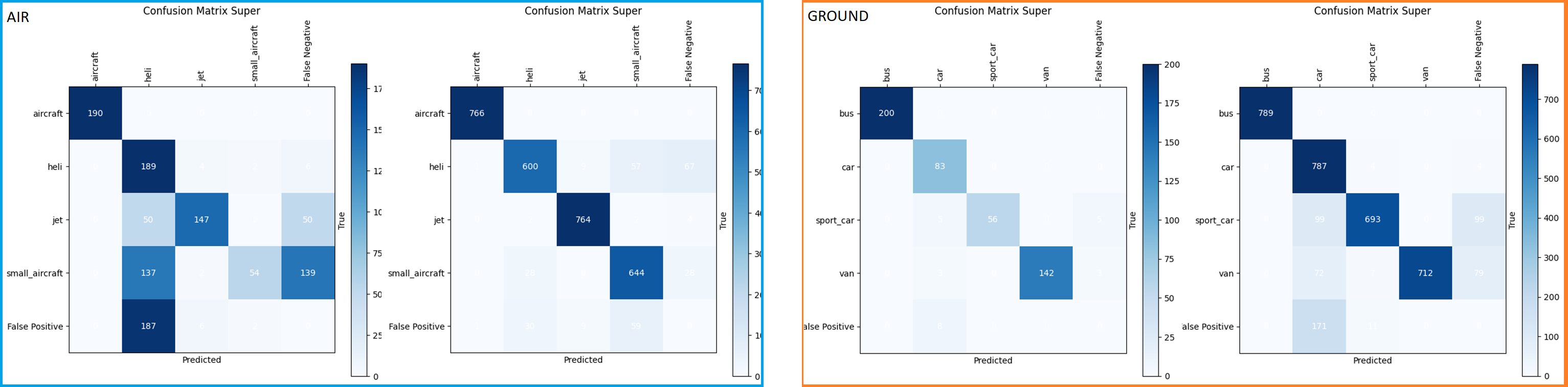}
 \end{center}
 \caption{Two comparisons between initial fine-tuning and extensive fine-tuning of Air and Ground categories. From left to right: the blue box is grouping Air category (initial and extensive correspondingly), and the orange box groups Ground category (initial and extensive correspondingly).}
 \label{fig:8}
\end{figure*}

The proposed methodology leverages a series of experiments \ref{table:experiment_specs} to comprehensively assess its performance. In the initial experiment, the model was subjected to 10 epochs of training with a batch size of 3, utilizing synthetic data that spanned four distinct categories. During this training phase, the learning rate was set to 0.0001, and the optimization process was facilitated by the Adam optimizer. Importantly, data augmentation techniques were not employed during this phase.

After the initial experiments detailed above, additional experiments were carried out to probe the impact of extended training on the performance of the models. These final experiments didn't involve the utilization of fine-tuned models that had undergone 10 epochs of training but were done from the beginning. The choice of batch size for these experiments varied from 3 to 6, contingent on the memory limitations of the GPU.

Furthermore, to expand the scope and depth of the analysis, extended experiments were conducted. These experiments employed the same fine-tuned models, which had undergone 100 epochs of training, and maintained similar batch size ranges as the final experiments. This extended training duration aimed to provide a more comprehensive assessment of the models' performance, encompassing a broader range of training scenarios and conditions.

\begin{table*}[!htb]
\small
\center
\caption{Parameter Tuning}
\label{table:experiment_specs}
\resizebox{1\textwidth}{!}{%
\begin{tabular}{|l|l|l|l|l|l|l|l|}
\hline
\textbf{Experiment Phase} & \textbf{Epochs} & \textbf{Batch Size} & \textbf{Data Type} & \textbf{Learning Rate} & \textbf{Optimizer} & \textbf{Synth. Data} & \textbf{Notes} \\ \hline
Initial         & 10                       & 3                   & 4 categories & 0.0001              & Adam               & $\times$              & -                         \\ \hline
Additional    & 10                       & 3-6  & 4 categories                      & 0.0001                    & Adam                   & $\checkmark$                         & GPU memory limitations   \\ \hline
Extended      & 100                      & 3-6    & 8 categories                      & 0.0001                    & Adam                  &  $\checkmark$                        & Fine-tuning, performance assessment \\ \hline
\end{tabular}
}

\end{table*}

% Text is done, needs diagrams.
\section{Evaluation}
\label{Evaluation}
This section aims to showcase the results and evaluation of the proposed methodology, providing a comparative study with similar benchmark methods. The experiments performed, data and quantitative metrics used are described below.

\subsection{Metrics}
The evaluation of the proposed method relied on the mean Average Precision (mAP), a standard metric to quantify object detection performance based on a user-defined set of criteria \cite{lin2014microsoft}. It is defined as the mean value of the average precision of the individual classes:

\begin{equation}
    mAP=\frac{1}{n}\sum_{k=1}^{n}AP_{k}
\end{equation}
where \(AP_{k}\) is Average Precision of class $k$, and $n$ is the number of classes.

An extensive evaluation has been conducted using a synthetic dataset that we created using a 3D Rendering Engine. This dataset consists of approximately 3000 images per category of vehicles in different environments and weather conditions. An example of the dataset with predictions can be seen in Figure \ref{fig:7}. The images in this dataset belong to four different categories, each allowing us to assess our model on specific properties: a) Camera, b) Light, c) Weather, d) Sensor. The categories were assembled in such a way as to evaluate the model on specific properties and parameters of the virtual environment. Furthermore, each of the produced categories was split into Air and Ground subcategories where the Air subcategory contains only air vehicles, and the Ground one contains only ground vehicles.

\subsection{Data Specification \& Parameters}
As it was mentioned above, there are four categories of sub-datasets a) Camera, b) Light, c) Weather, and d) Sensor, Table \ref{table:data_specifications}. The Camera category represents images generated with different camera angles (point of view) and distances from an object for the City and the Desert scenes. Specifically, for the Air sub-category, there are images generated at 4 equal distances between 70 and 350 metres. For the Ground sub-category, there are images generated at 4 equal distances between 15 and 75 metres. In both categories the images were generated at 4 equal elevation angles between 5° and 85° degrees, and at 3 equal azimuth angles between 0° and 240° degrees. The other parameters such as light, image type, fog and rain were selected in such a way to prevent generating bias on the evaluation of the camera parameters.

The Light category contains images generated using variable balanced lighting parameters covering the City and Desert scenes. In more detail, the Air and Ground sub-categories were generated with the light intensity set between 10\% and 100\% power at 3 equal steps. The light elevation angles were set between 5° and 90° degrees at 3 equal steps, the light azimuth angles were set between 0° and 180° degrees at 3 equal steps. The other parameters related to camera, weather, and sensors were selected randomly and uniformly in such a way to avoid bias on the evaluation of the model under the set of the light parameters.

The Weather category contains images generated using different balanced weather parameters covering the City and Desert scenes. The Air and Ground sub-categories were generated both with and without enabling rain. Furthermore, the rainy images included variations due to the wind parameter that was selected to be 0 or 10 units of power. The other parameters were selected in such a way to avoid bias on the evaluation of the models in the weather category.

The night and thermal vision are the main attributes of the Sensor category. The night vision visualises an approximation of the effect of night vision goggles and the same approach was considered for the thermal vision. The Sensor category contains images generated using different balanced sensor image types covering the City and Desert scenes. Also, the Air and Ground sub-categories contain images emulating both night and thermal vision sensors. The other parameters again were selected uniformly preventing bias on the evaluation of the models for the sensor set of parameters.

\begin{table}[h]
\centering
\caption{Summary of Data Specifications}
\label{table:data_specifications}
\resizebox{1\columnwidth}{!}{%
\begin{tabular}{|p{2.7cm}|p{1cm}|p{2cm}|p{2cm}|p{2.2cm}|}
\hline
\textbf{Data Category} & \textbf{Scene} & \textbf{Sub-Categories} & \textbf{Parameters} & \textbf{Details} \\
\hline
Camera & City \& Desert & Air, Ground & Distance, Elevation Angle, Azimuth Angle & Distances: 70-350m (Air), 15-75m (Ground); Elevation Angles: 5°-85°; Azimuth Angles: 0°-240° \\
\hline
Light & City \& Desert & Air, Ground & Light Intensity, Elevation Angle, Azimuth Angle & Intensity: 10-100\% (3 steps); Elevation Angles: 5°-90°; Azimuth Angles: 0°-180° \\
\hline
Weather & City \& Desert & Air, Ground & Rain, Wind & Rain: Enabled/Disabled; Wind: 0 or 10 units \\
\hline
Sensor & City \& Desert & Air, Ground & Night Vision, Thermal Vision & Night and Thermal Vision emulations \\
\hline
Real (KITTI) & Varied & Air, Ground & High-resolution Images, Lidar, Calibration & Object Detection, Tracking, 3D Scene Understanding \\
\hline
\end{tabular}
}
\end{table}

\subsection{Real Dataset Experiments}

The Real dataset was the KITTI dataset \cite{Geiger2012CVPR} which is a widely used benchmark dataset for research in computer vision and autonomous driving \cite{Mao2023}. It stands for "Karlsruhe Institute of Technology and Toyota Technological Institute" and was created by researchers from these institutions. This dataset is commonly referenced in academic publications related to tasks such as object detection, tracking, 3D scene understanding, and more. The specification of the dataset can be seen in Table \ref{table:kitti}.

\begin{figure}
 \begin{center}   
  \includegraphics[width=0.9\textwidth]{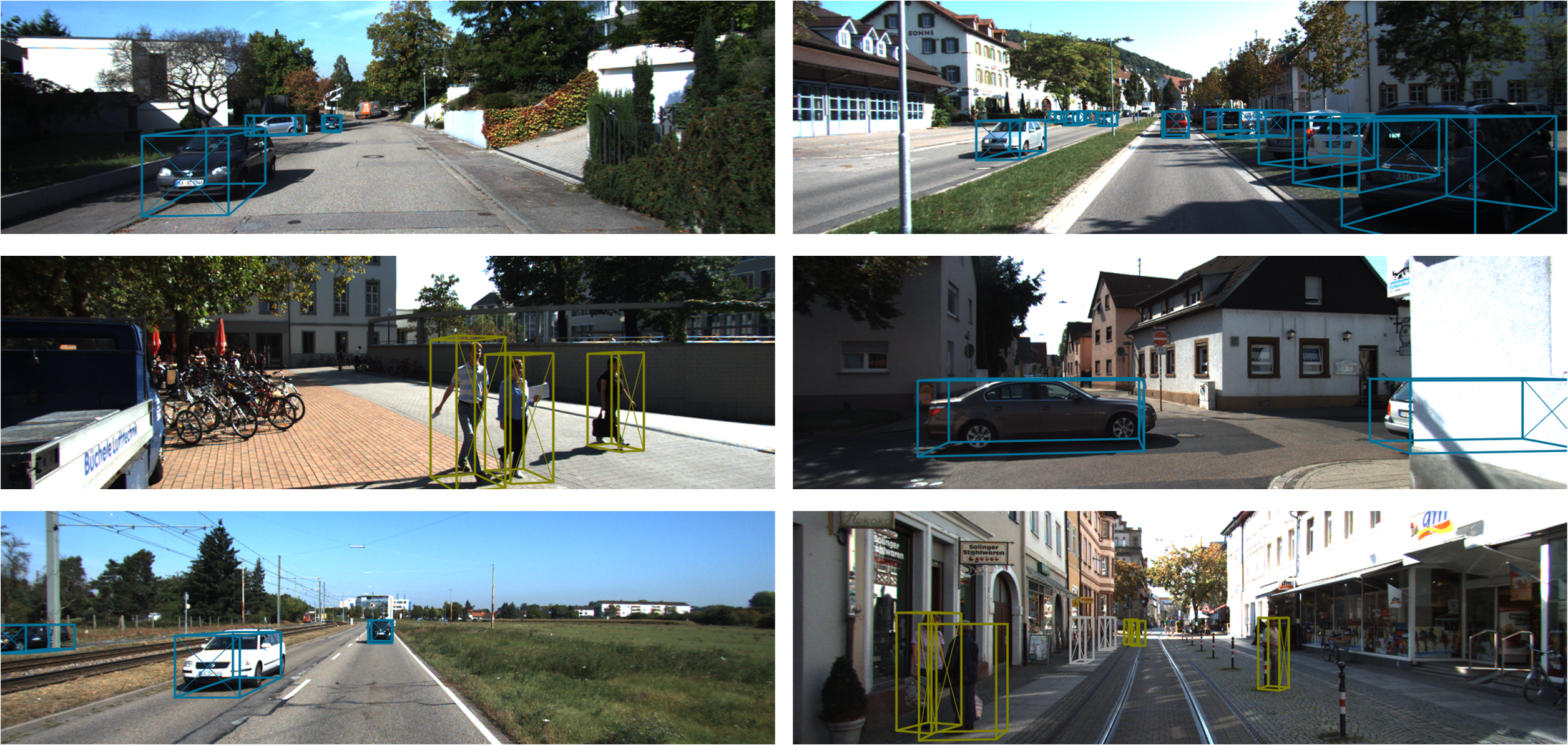}
 \end{center}
 \caption{Real Dataset 3D Bounding Box Sample}
 \label{fig:10}
\end{figure}

\begin{table}[h]
\centering
\caption{Features and Specifications of the 3D Object Detection KITTI Dataset}
\label{table:kitti}
\resizebox{1\columnwidth}{!}{%
\begin{tabular}{|p{4cm}|p{7cm}|}
\hline
\textbf{Feature/Specification}    & \textbf{Description}                                                                 \\ \hline
\textbf{Data Type}                & Images, Lidar data                                                                  \\ \hline
\textbf{Tasks}                    & Stereo, Optical Flow, Visual Odometry, 3D Object Detection, Tracking                 \\ \hline
\textbf{Number of Images}         & \textasciitilde15,000 images for object detection                                   \\ \hline
\textbf{Image Resolution}         & 1242 x 375 pixels                                                                   \\ \hline
\textbf{Sensors}  & Inertial Navigation System (GPS/IMU): OXTS RT 3003, Laserscanner: Velodyne HDL-64E, Grayscale cameras, 1.4 Megapixels: Point Grey Flea 2 (FL2-14S3M-C), Color cameras, 1.4 Megapixels: Point Grey Flea 2 (FL2-14S3C-C), Varifocal lenses, 4-8 mm: Edmund Optics NT59-917                                                       \\ \hline
\textbf{Annotation Types}         & Bounding boxes, 3D boxes, object type, truncation, occlusion levels                 \\ \hline
\textbf{Environments}             & Urban, residential, road                                                            \\ \hline
\textbf{Classes}                  & Cars, vans, trucks, pedestrians, cyclists                                           \\ \hline
\textbf{Ground Truth Availability}& Yes                                                              \\ \hline
\end{tabular}
}
\end{table}

This dataset under consideration in this study is the KITTI dataset, an acronym for "Karlsruhe Institute of Technology and Toyota Technological Institute." This benchmark dataset is widely used within the domain of computer vision and autonomous driving research. The primary objective behind its inception is to foster the advancement of algorithms and technologies relevant to autonomous vehicles. The dataset is characterized by a comprehensive collection of diverse data modalities, encompassing high-resolution camera images, lidar point clouds, and calibration parameters. This dataset is used for a wide variety of tasks, including object detection, motion tracking, 3D scene analysis, and other such applications. An added feature of the KITTI dataset is the provision of image annotations for various object types, such as cars, pedestrians, and cyclists, thereby rendering it an important resource for the validation of AI models. Furthermore, the dataset encompasses a wide spectrum of real-world driving scenarios, variable weather conditions, and different times of day, thereby facilitating a comprehensive assessment of algorithm performance under diverse environmental conditions. It is worth noting that while the KITTI dataset is widely used in the research community, it does exhibit certain limitations, notably its relatively modest scale and the absence of data pertaining to certain object classes, e.g., motorcycles.

The performance of the proposed framework for the aforementioned four categories using mAP could be seen in the Table \ref{tab:synthetic-results} and \ref{tab:real-results}. Additionally, the confusion matrices can be observed in Figure 
\ref{fig:8}. 

% \begin{table}[]
%  \centering
%   \begin{tabular}{ | c | c | c | c | }
%    \hline
%    Category & Sub-category  & 10 epochs & 100 epochs \\\hline

%    \multirow{4}{*}{Air} & Camera  &         30.00\%  & \textbf{61.04\%} \\\cline{2-4}
%                         & Light   & \textbf{39.95\%} &         34.61\%  \\\cline{2-4}
%                         & Weather &         35.83\%  & \textbf{88.71\%} \\\cline{2-4}
%                         & Sensor  &         32.51\%  & \textbf{51.90\%} \\\hline
%    \multirow{4}{*}{Ground} & Camera  & \textbf{74.66\%} &         61.02\%  \\\cline{2-4}
%                            & Light   &         25.05\%  & \textbf{33.82\%} \\\cline{2-4}
%                            & Weather &         23.45\%  & \textbf{58.75\%} \\\cline{2-4}
%                            & Sensor  &         36.67\%  & \textbf{55.72\%} \\\hline
%   \end{tabular}
%  \caption{Results of the experiments on the Synthetic dataset (mAP)}
%  \label{tab:synthetic-results}
% \end{table}

\begin{table}[]
 \centering
 \caption{Performance results on the Synthetic dataset (mAP)}
 \label{tab:synthetic-results}
 \resizebox{1\columnwidth}{!}{%
  \begin{tabular}{ | c | c | c | c | c | c | }
   \hline
   Category & Sub-category  & CenterNet & FRRCNN  & YOLOv3  & RETINA \\\hline

   \multirow{4}{*}{Air} & Camera  & \textbf{61.04\%} & 5.24\%  & 44.82\% & 44.79\% \\\cline{2-6}
                        & Light   & \textbf{39.95\%} & 20.66\% & 63.58\% & 61.25\% \\\cline{2-6}
                        & Weather & \textbf{88.71\%} & 5.35\%  & 39.00\% & 45.57\% \\\cline{2-6}
                        & Sensor  & \textbf{51.90\%} & 4.97\%  & 4.27\%  & 7.95\%  \\\hline
   \multirow{4}{*}{Ground} & Camera  & \textbf{74.66\%} & 17.95\% & 76.02\% & 88.81\% \\\cline{2-6}
                           & Light   & \textbf{33.82\%} & 32.54\% & 38.52\% & 87.12\% \\\cline{2-6}
                           & Weather & \textbf{58.75\%} & 17.07\% & 66.32\% & 86.09\% \\\cline{2-6}
                           & Sensor  & \textbf{55.72\%} & 4.16\%  & 7.64\%  & 15.59\% \\\hline
  \end{tabular}
 }
\end{table}

\begin{table}[]
 \centering
  \caption{Performance results on the Real dataset (mAP)}
 \label{tab:real-results}
  \begin{tabular}{ | l | c | }
   \hline
      \textbf{Class}      & \textbf{\%}        \\
   \hline
      Car        & 87.846443 \\\hline
      Pedestrian & 60.852219 \\\hline
      Cyclist    & 48.693352 \\
   \hline
      \multicolumn{2}{ | c | }{\textbf{Total}} \\
   \hline
      mAP & \textbf{65.797338} \\
   \hline
  \end{tabular}

\end{table}

The presented methodology found practical application in the domains of object recognition and scene comprehension. An evaluation of its performance was carried out using a synthetic dataset intentionally designed to introduce diverse environmental conditions that would influence image quality as well as using a dataset consisting out of images from real environments. In the ensuing discussion, we outline the detailed analysis of the evaluation conducted on the CenterNet model. In summary, the CenterNet model exhibited a consistent level of stability and predictability, denoting its ability to deliver reliable results across varying environmental conditions.

\begin{figure}
 \begin{center}   
  \includegraphics[width=0.9\textwidth]{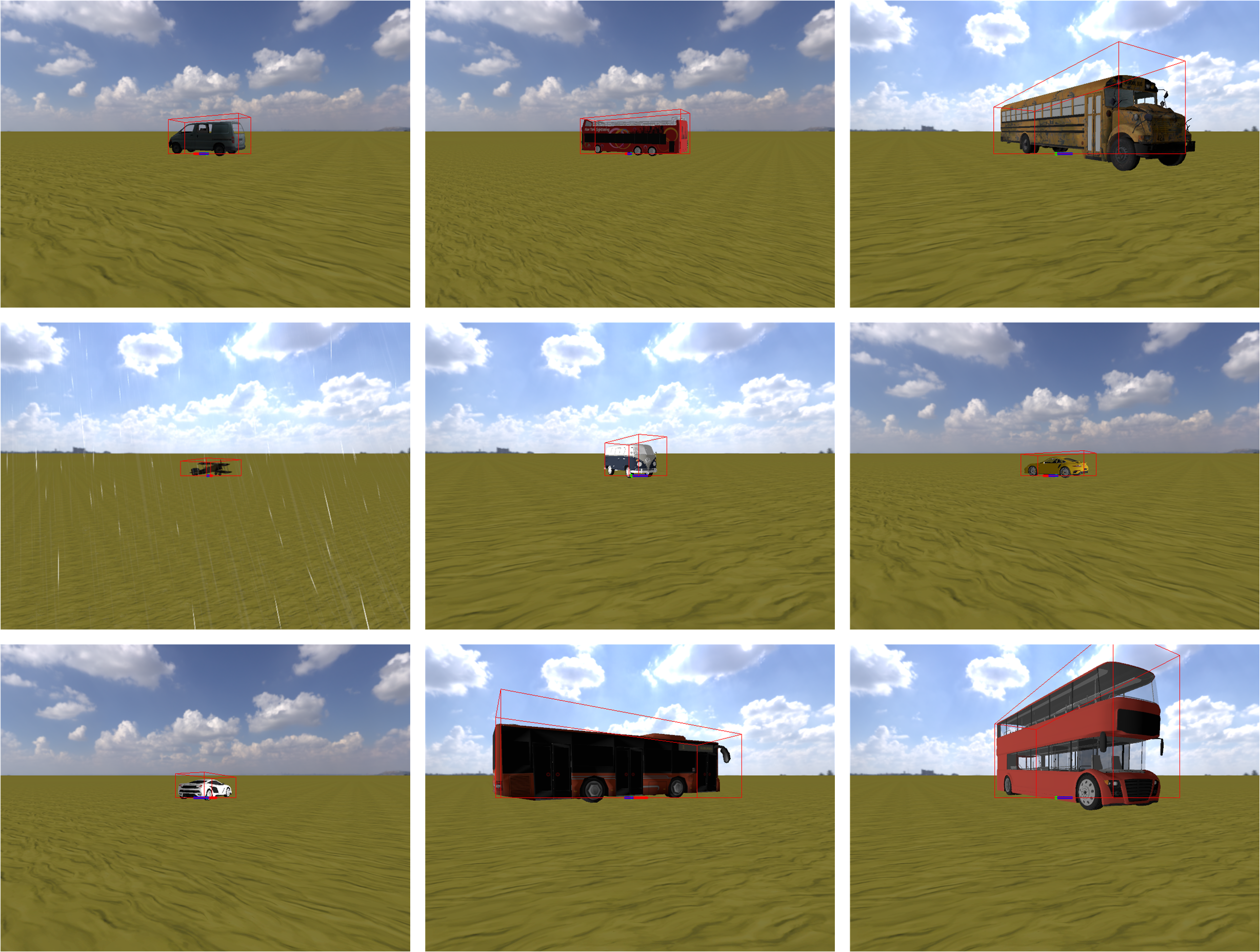}
 \end{center}
 \caption{Example of 3D bounding box predictions. (Synthetic Data)}
 \label{fig:10}
\end{figure}

The results of the Table \ref{tab:synthetic-results} seemed to show competitive results comparing it with the results of the Table \ref{tab:real-results}. Table \ref{tab:real-results} demonstrating state of the art performance with a high precision in the "Car" category. As the real dataset was is unbalanced towards the "Car" category the "Pedestrian" and "Cyclist" categories demonstrate worse results. The "Cyclist" category is the smallest category in the real dataset and consequently performed the worst. Additionally, visual similarity between "Pedestrians" and "Cyclists" somewhat contributes towards decreased results.

Table \ref{tab:synthetic-results} displays the results obtained by CenterNet on the Air and Ground datasets. Figure \ref{fig:6} shows the confusion matrices of the CenterNet model between the Air and Ground datasets. The CenterNet model produced competitive results for the Air sub-category. When comparing the Air and Ground results the ground vehicles were detected with significantly higher accuracy. The most likely cause of such behaviour is related with the attributes of the Air and Ground datasets. The main difference between the subsets is the supported distances between the camera and the object. Furthermore, the objects themselves are more diverse in terms of size and shape.

As per the results in Table \ref{tab:synthetic-results} show, the easiest category to detect was the Light category. In both sub-categories performance was relatively high taking into consideration the number of epochs used for the training. The Light sub-category includes images with similar parameters and visual features, for example, multiple images of the objects at certain angles. According to the results, the Weather subset results ended up around the middle of the output performance table. Comparing Air and Ground results, the model performed well with a general trend of increased accuracy for the Air sub-category.

The fourth subset, the Sensor subset, generated for this project was focused on the special types of sensors. Examples of predictions are illustrated in Figure \ref{fig:7} for the CenterNet models. Analysing Table \ref{tab:synthetic-results}, the Sensor subset placed itself as the most challenging for the CenterNet model.

In the subsequent stage of experimentation, the ML architecture underwent fine-tuning, involving a rigorous training regimen spanning 100 epochs. This fine-tuning process was executed separately for each of the primary categories, namely "Air" and "Ground," as well as for each of the sub-categories, namely "Camera," "Light," "Weather," and "Sensor."

\subsection{Synthetic Dataset Additional Experiments}

For the task of extended evaluation, two distinct sets of data were amalgamated from the comprehensive Synthetic dataset, utilizing identical parameters. The first dataset encompassed images captured in the "Forest" and "Grass" scenes, while the second dataset comprised images originating from the "City" and "Desert" scenes. Notably, each of these datasets comprised approximately 3000 images for each sub-category.

The allocation of images into training, validation, and testing subsets adhered to a specific protocol. The training subset exclusively consisted of images derived from the "Forest" and "Grass" scenes, encompassing 100\% of the images within the first dataset. The validation subset, on the other hand, was composed of a tenth of all images from the "City" and "Desert" scenes, representing 10\% of the images in the second dataset. The remaining images from the "City" and "Desert" scenes were collectively grouped into the testing subset, thereby establishing a structured and well-defined partitioning of the data for the purposes of robust evaluation.

The results of the experiments demonstrate the performance of the new ML architecture via a mean average precision (mAP) in percentages. In comparison with the final experiments, the extended experiments, depicted, demonstrated an overall performance increase.

Regarding the "Air" category, a noticeable gain could be observed in the Camera sub-category. On the other hand, the "Light" sub-category has minor deviations. However, the results of the "Light" sub-category were already high, therefore dramatic changes weren’t expected. In the "Weather" sub-category similar to the "Camera" sub-category, the improvement of the CenterNet model is noticeable. Although, the type of images in the "Sensor" sub-category differs from the types of images in the other sub-categories, all the models reached better performance metrics.

Switching to the "Ground" category, covering the same four sub-categories: "Camera", "Light", "Weather", and "Sensor".The "Camera" sub-category had high results in the final experiments, the extended experiments produced slightly deviating results. The "Light" sub-category followed similar pattern where the final experiments produced moderately better results in the extended experiments. The CenterNet model had some room for improvement therefore the positive change was more noticeable. The "Weather" sub-category continued the trend. The CenterNet model showed considerable increase. The "Sensor" sub-category was akin to the "Sensor" sub-category under the "Air" category where concordant improvement was seen. The results suggest that the CenterNet model benefited from the extensive training and gained an improvement.

\FloatBarrier

% DONE
\section{Conclusion}
\label{Conclusion}
This paper provides a comprehensive overview of existing methodologies and approaches within the realm of scene analysis, leveraged by autonomous vehicles, with a specific emphasis on their applicability in immersive environments. The research presented delves into an in-depth analysis of a 3D object detection model from the vantage point of the augmented reality domain. The architectural framework comprises a diverse set of components, each meticulously designed to tackle various intricacies related to the estimation of keypoints, the conversion of keypoints to 2D bounding boxes, and the inference of crucial spatial information. This information encompasses depth, 3D dimensions measured in meters, as well as orientation, encompassing azimuth, elevation, and roll angles. The collective contributions of these components culminate in a model that exhibits proficiency in the projection of 3D bounding boxes onto a 2D image.

To empirically evaluate the efficacy of the proposed architecture, a comprehensive testing regimen was conducted, utilizing a synthetic dataset in a comparative study. The outcomes of this evaluation reveal that the proposed model delivers competitive performance while demonstrating stability, particularly when tasked with the detection of distant objects. The evaluation and analysis of the proposed model were undertaken under diverse environmental conditions and with varying camera settings, establishing its versatility and robustness. Furthermore, to augment the comprehensiveness of the study, a novel and well-balanced synthetic dataset was created and curated, utilising a virtual environment. This dataset encompasses annotated data spanning a multitude of objects and environmental scenarios, providing a rich resource for subsequent validation, experimentation, and refinement.

\section*{Acknowledgments}
This project has received funding from the European Union’s Horizon Europe research and innovation programme under grant agreement No. 101070181 (TALON).

\bibliographystyle{elsarticle-num} 
\bibliography{main}

\begin{thebibliography}{10}
\expandafter\ifx\csname url\endcsname\relax
  \def\url#1{\texttt{#1}}\fi
\expandafter\ifx\csname urlprefix\endcsname\relax\def\urlprefix{URL }\fi
\expandafter\ifx\csname href\endcsname\relax
  \def\href#1#2{#2} \def\path#1{#1}\fi

\bibitem{pavel2022vision}
M.~I. Pavel, S.~Y. Tan, A.~Abdullah, Vision-based autonomous vehicle systems based on deep learning: A systematic literature review, Applied Sciences 12~(14) (2022) 6831.

\bibitem{xiong2021augmented}
J.~Xiong, E.-L. Hsiang, Z.~He, T.~Zhan, S.-T. Wu, Augmented reality and virtual reality displays: emerging technologies and future perspectives, Light: Science \& Applications 10~(1) (2021) 216.

\bibitem{zou2023object}
Z.~Zou, K.~Chen, Z.~Shi, Y.~Guo, J.~Ye, Object detection in 20 years: A survey, Proceedings of the IEEE (2023).

\bibitem{anderson2019feasibility}
R.~Anderson, J.~Toledo, H.~ElAarag, Feasibility study on the utilization of microsoft hololens to increase driving conditions awareness, in: 2019 SoutheastCon, IEEE, 2019, pp. 1--8.

\bibitem{gomes2018augmented}
D.~L. Gomes~Jr, A.~C. de~Paiva, A.~C. Silva, G.~Braz~Jr, J.~D.~S. de~Almeida, A.~S. de~Ara{\'u}jo, M.~Gattas, Augmented visualization using homomorphic filtering and haar-based natural markers for power systems substations, Computers in Industry 97 (2018) 67--75.

\bibitem{dimitropoulos2021operator}
N.~Dimitropoulos, T.~Togias, G.~Michalos, S.~Makris, Operator support in human--robot collaborative environments using ai enhanced wearable devices, Procedia Cirp 97 (2021) 464--469.

\bibitem{zhao2019object}
Z.-Q. Zhao, P.~Zheng, S.-t. Xu, X.~Wu, Object detection with deep learning: A review, IEEE transactions on neural networks and learning systems 30~(11) (2019) 3212--3232.

\bibitem{lecun1998gradient}
Y.~LeCun, L.~Bottou, Y.~Bengio, P.~Haffner, Gradient-based learning applied to document recognition, Proceedings of the IEEE 86~(11) (1998) 2278--2324.

\bibitem{girshick2014rich}
R.~Girshick, J.~Donahue, T.~Darrell, J.~Malik, Rich feature hierarchies for accurate object detection and semantic segmentation, in: Proceedings of the IEEE conference on computer vision and pattern recognition, 2014, pp. 580--587.

\bibitem{viola2001rapid}
P.~Viola, M.~Jones, Rapid object detection using a boosted cascade of simple features, in: Proceedings of the 2001 IEEE computer society conference on computer vision and pattern recognition. CVPR 2001, Vol.~1, Ieee, 2001, pp. I--I.

\bibitem{he2015spatial}
K.~He, X.~Zhang, S.~Ren, J.~Sun, Spatial pyramid pooling in deep convolutional networks for visual recognition, IEEE transactions on pattern analysis and machine intelligence 37~(9) (2015) 1904--1916.

\bibitem{girshick2015fast}
R.~Girshick, Fast r-cnn, in: Proceedings of the IEEE international conference on computer vision, 2015, pp. 1440--1448.

\bibitem{ren2015faster}
S.~Ren, K.~He, R.~Girshick, J.~Sun, Faster r-cnn: Towards real-time object detection with region proposal networks, Advances in neural information processing systems 28 (2015).

\bibitem{dai2016r}
J.~Dai, Y.~Li, K.~He, J.~Sun, R-fcn: Object detection via region-based fully convolutional networks, Advances in neural information processing systems 29 (2016).

\bibitem{li2017light}
Z.~Li, C.~Peng, G.~Yu, X.~Zhang, Y.~Deng, J.~Sun, Light-head r-cnn: In defense of two-stage object detector, arXiv preprint arXiv:1711.07264 (2017).

\bibitem{lin2017feature}
T.-Y. Lin, P.~Doll{\'a}r, R.~Girshick, K.~He, B.~Hariharan, S.~Belongie, Feature pyramid networks for object detection, in: Proceedings of the IEEE conference on computer vision and pattern recognition, 2017, pp. 2117--2125.

\bibitem{cao2020d2det}
J.~Cao, H.~Cholakkal, R.~M. Anwer, F.~S. Khan, Y.~Pang, L.~Shao, D2det: Towards high quality object detection and instance segmentation, in: Proceedings of the IEEE/CVF conference on computer vision and pattern recognition, 2020, pp. 11485--11494.

\bibitem{redmon2016you}
J.~Redmon, S.~Divvala, R.~Girshick, A.~Farhadi, You only look once: Unified, real-time object detection, in: Proceedings of the IEEE conference on computer vision and pattern recognition, 2016, pp. 779--788.

\bibitem{redmon2017yolo9000}
J.~Redmon, A.~Farhadi, Yolo9000: better, faster, stronger, in: Proceedings of the IEEE conference on computer vision and pattern recognition, 2017, pp. 7263--7271.

\bibitem{redmon2018yolov3}
J.~Redmon, A.~Farhadi, Yolov3: An incremental improvement, arXiv preprint arXiv:1804.02767 (2018).

\bibitem{Wang2022}
Z.~Wang, L.~Wu, T.~Li, P.~Shi, \href{https://www.mdpi.com/2227-7390/10/7/1190}{A smoke detection model based on improved yolov5}, Mathematics 10~(7) (2022).
\newblock \href {https://doi.org/10.3390/math10071190} {\path{doi:10.3390/math10071190}}.
\newline\urlprefix\url{https://www.mdpi.com/2227-7390/10/7/1190}

\bibitem{liu2016ssd}
W.~Liu, D.~Anguelov, D.~Erhan, C.~Szegedy, S.~Reed, C.-Y. Fu, A.~C. Berg, Ssd: Single shot multibox detector, in: Computer Vision--ECCV 2016: 14th European Conference, Amsterdam, The Netherlands, October 11--14, 2016, Proceedings, Part I 14, Springer, 2016, pp. 21--37.

\bibitem{lin2017focal}
T.-Y. Lin, P.~Goyal, R.~Girshick, K.~He, P.~Doll{\'a}r, Focal loss for dense object detection, in: Proceedings of the IEEE international conference on computer vision, 2017, pp. 2980--2988.

\bibitem{Xinyu2023}
X.~Wu, D.~Ma, X.~Qu, X.~Jiang, D.~Zeng, \href{https://www.sciencedirect.com/science/article/pii/S092523122201414X}{Depth dynamic center difference convolutions for monocular 3d object detection}, Neurocomputing 520 (2023) 73--81.
\newblock \href {https://doi.org/https://doi.org/10.1016/j.neucom.2022.11.032} {\path{doi:https://doi.org/10.1016/j.neucom.2022.11.032}}.
\newline\urlprefix\url{https://www.sciencedirect.com/science/article/pii/S092523122201414X}

\bibitem{zhou2019objects}
X.~Zhou, D.~Wang, P.~Kr{\"a}henb{\"u}hl, Objects as points, arXiv preprint arXiv:1904.07850 (2019).

\bibitem{law2018cornernet}
H.~Law, J.~Deng, Cornernet: Detecting objects as paired keypoints, in: Proceedings of the European conference on computer vision (ECCV), 2018, pp. 734--750.

\bibitem{Perez_Rua_2020_CVPR}
J.-M. Perez-Rua, X.~Zhu, T.~M. Hospedales, T.~Xiang, Incremental few-shot object detection, in: Proceedings of the IEEE/CVF Conference on Computer Vision and Pattern Recognition (CVPR), 2020.

\bibitem{carion2020end}
N.~Carion, F.~Massa, G.~Synnaeve, N.~Usunier, A.~Kirillov, S.~Zagoruyko, End-to-end object detection with transformers, in: Computer Vision--ECCV 2020: 16th European Conference, Glasgow, UK, August 23--28, 2020, Proceedings, Part I 16, Springer, 2020, pp. 213--229.

\bibitem{zhu2021deformable}
X.~Zhu, W.~Su, L.~Lu, B.~Li, X.~Wang, J.~Dai, Deformable detr: Deformable transformers for end-to-end object detection (2021).
\newblock \href {http://arxiv.org/abs/2010.04159} {\path{arXiv:2010.04159}}.

\bibitem{duan2019centernet}
K.~Duan, S.~Bai, L.~Xie, H.~Qi, Q.~Huang, Q.~Tian, Centernet: Keypoint triplets for object detection, in: Proceedings of the IEEE/CVF international conference on computer vision, 2019, pp. 6569--6578.

\bibitem{eigen2014depth}
D.~Eigen, C.~Puhrsch, R.~Fergus, Depth map prediction from a single image using a multi-scale deep network, Advances in neural information processing systems 27 (2014).

\bibitem{mousavian20173d}
A.~Mousavian, D.~Anguelov, J.~Flynn, J.~Kosecka, 3d bounding box estimation using deep learning and geometry, in: Proceedings of the IEEE conference on Computer Vision and Pattern Recognition, 2017, pp. 7074--7082.

\bibitem{lin2014microsoft}
T.-Y. Lin, M.~Maire, S.~Belongie, J.~Hays, P.~Perona, D.~Ramanan, P.~Doll{\'a}r, C.~L. Zitnick, Microsoft coco: Common objects in context, in: Computer Vision--ECCV 2014: 13th European Conference, Zurich, Switzerland, September 6-12, 2014, Proceedings, Part V 13, Springer, 2014, pp. 740--755.

\bibitem{Geiger2012CVPR}
A.~Geiger, P.~Lenz, R.~Urtasun, Are we ready for autonomous driving? the kitti vision benchmark suite, in: Conference on Computer Vision and Pattern Recognition (CVPR), 2012.

\bibitem{Mao2023}
J.~Mao, S.~Shi, X.~Wang, H.~Li, 3d object detection for autonomous driving: A comprehensive survey, International Journal of Computer Vision 131 (2023) 1--55.
\newblock \href {https://doi.org/10.1007/s11263-023-01790-1} {\path{doi:10.1007/s11263-023-01790-1}}.

\end{thebibliography}

\end{document}